\documentclass[10pt, twocolumn,letterpaper]{article}
\usepackage{wacv}
\usepackage{times}
\usepackage{epsfig}
\usepackage{graphicx}
\usepackage{amsmath}
\usepackage{amssymb}
\usepackage{multirow}
\usepackage{adjustbox}
\usepackage{subcaption}
\usepackage{pifont}
\usepackage[english]{babel}
\usepackage[utf8]{inputenc}
\usepackage{algorithm}
\usepackage[noend]{algpseudocode}

\usepackage[pagebackref=true,breaklinks=true,letterpaper=true,colorlinks,bookmarks=false]{hyperref}

\wacvfinalcopy 

\def\wacvPaperID{751} 

\ifwacvfinal\fi
\begin{document}

\title{Ortho-Shot: Low Displacement Rank Regularization with Data Augmentation for Few-Shot Learning
}

\author{Uche Osahor \\
West Virginia University\\
{\tt\small uo0002@mix.wvu.edu}
\and
Nasser M. Nasrabadi \\
West Virginia University\\
{\tt\small nasser.nasrabadi@mail.wvu.edu}
}

\maketitle
\pagestyle{plain}
\begin{abstract}
\vspace{-0.2cm}
In few-shot classification, the primary goal is to learn representations from a few samples that generalize well for novel classes. In this paper, we propose an efficient low displacement rank (LDR) regularization strategy termed Ortho-Shot; a technique that imposes orthogonal regularization  on the convolutional layers of a few-shot classifier, which  is based on the doubly-block toeplitz (DBT) matrix structure. The regularized convolutional layers of the few-shot classifier enhances model generalization and intra-class feature embeddings that are crucial for few-shot learning. Overfitting is a typical issue for few-shot models, the lack of data diversity inhibits proper model inference which weakens the classification accuracy of few-shot learners to novel classes. In this regard, we broke down the pipeline of the few-shot classifier and established that the support, query and task data augmentation collectively alleviates  overfitting in networks. With compelling results, we demonstrated that combining a DBT-based low-rank orthogonal regularizer with data augmentation strategies, significantly boosts the performance of a few-shot classifier. We perform our experiments on the miniImagenet, CIFAR-FS and Stanford datasets with performance values of about 5\% when compared to  state-of-the-art.
\vspace{-0.10cm}
\end{abstract}

\vspace{-.5cm}
\section{Introduction}
\vspace{-0.14cm}
The performance of convolutional neural network (CNN) models largely depend on training a network with a lot of labelled instances and a spectrum of visual variations which are mostly in thousands per class \cite{Krizhevsky2012ImageNetCW}. The cost of labelling these data manually by human annotation as well as the scarcity of data that captures the complete diversity in a specific class significantly limits the potential of current vision models. However, the human visual system (HVS) can identify new classes with fewer labelled examples \cite{Kietzmann2019RecurrenceIR, Nayebi2018TaskDrivenCR}, this unique trait of the HVS reveals the need to dive into new paradigms that would learn to generalize new classes with a limited amount of labelled data for each novel class. Recently, significant progress  has been made towards better solutions using ideas of meta-learning \cite{Oreshkin2018TADAMTD, Rusu2019MetaLearningWL, Ye2018LearningEA, Lee2019MetaLearningWD, Li2019FindingTF, Motiian2017FewShotAD}.
\begin{figure}[t!]
\begin{center}
  \includegraphics[width=8.3cm, height=5cm]{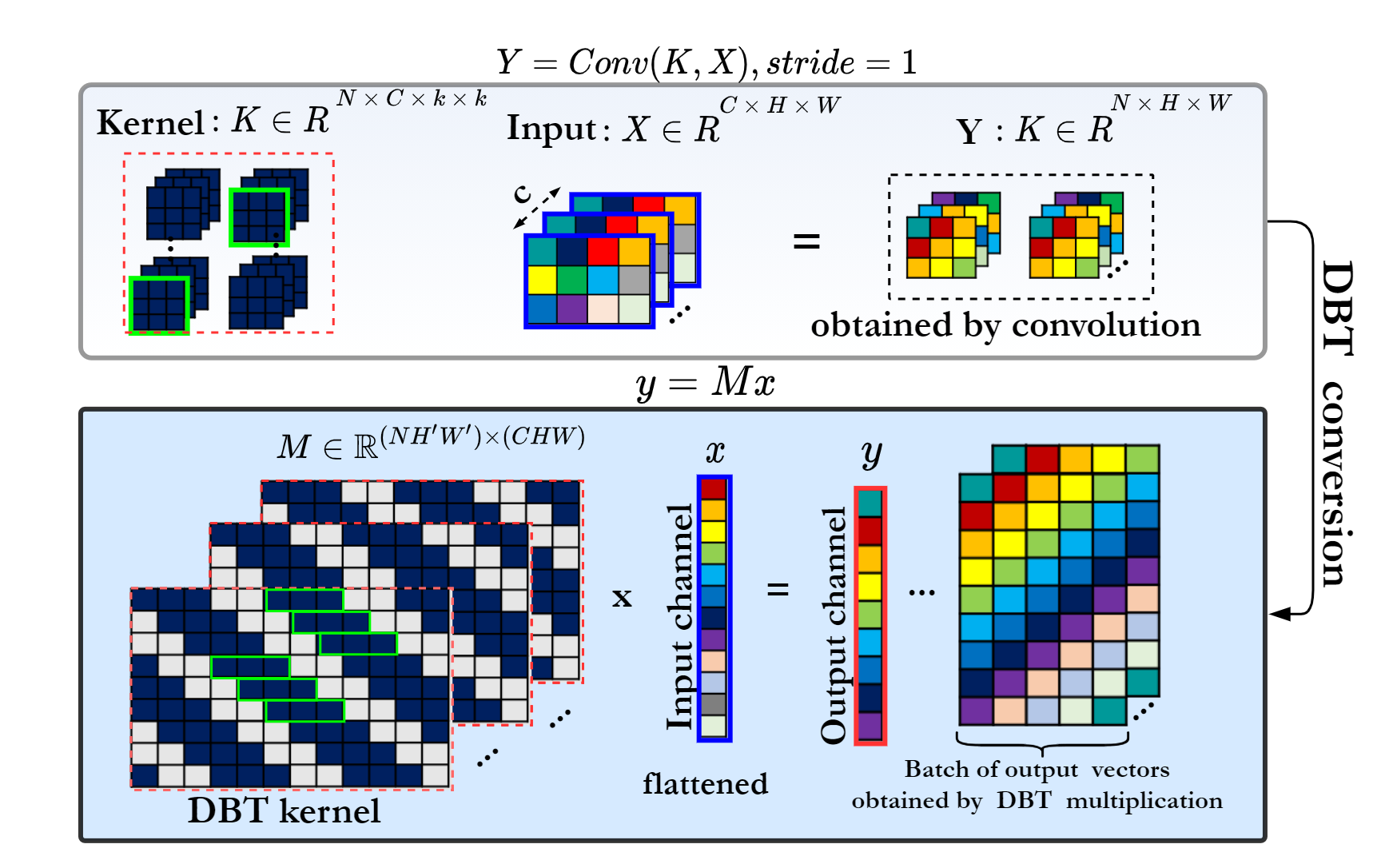}
\end{center}
\vspace{-0.5cm}
  \caption{The convolution expression; Conv(K, X) is converted into a faster DBT vector representation; $ y = Mx$\label{orthoT}.
}
\vspace{-0.5cm}
\end{figure}
Empirically, it has been observed that the convolutional filters learned in deeper layers are highly correlated and redundant \cite{Wang2020OrthogonalCN}, thereby resulting in unstable training performance and vanishing gradients.  These shortcomings of convolutional neural networks are also more damaging in few-shot classification due to the small data size. The potential pitfalls of such convolutional layers could result in under-utilization of model capacity, overfitting, vanishing and exploding gradients \cite{Glorot2010UnderstandingTD, Bengio1994LearningLD}, growth in saddle points \cite{Dauphin2014IdentifyingAA} and shifts in feature statistics \cite{Ioffe2015BatchNA}, which collectively affect model generalization. 

The doubly block-toeplitz (DBT) matrix \cite{Gray2005ToeplitzAC} is part of a class of low displacement rank (LDR) matrix constructions \cite{Zhao2017TheoreticalPF} that guarantee model reduction and computational complexity reduction in neural networks which is achieved by regularizing the weight matrices of network layers. The storage requirement of such a DBT-regularized network is reduced from $O(n^2)$  to $O(n)$ and the computational complexity can be reduced from $O(n^2)$ to  $ O(nrlogn)$, due to the fast matrix-vector multiplication property of LDR structured matrices as shown in Figure 2. It is also well established \cite{Li2017LowRankDE,Thomas2018LearningCT} that when filters are learned to be as orthogonal as possible, model capacity is better utilized which in-turn improves feature expressiveness and intra-class feature representation \cite{Araujo2020OnLR,Vapnik2000TheNO,thomas2019learning, Aghdaie2021AttentionAW}. 

Our goal is to present an effective baseline model that harnesses good learned representations for few-shot classification kinds of tasks  which perform better or at par with current few-shot algorithms \cite{Wang2016LearningTL, Vinyals2016MatchingNF, Triantafillou2017FewShotLT, Snell2017PrototypicalNF, Sung2018LearningTC, Oreshkin2018TADAMTD, Rusu2019MetaLearningWL, Motiian2017FewShotAD}. In a nutshell, we tackled the few-shot learning limitations by imposing orthogonal regularization on the model baseline which is a simpler yet effective approach compared to techniques used previously in \cite{Wang2018LowShotLF, Gidaris2018DynamicFV, Qi2018LowShotLW}. We also incorporated data  augmentation  strategies  that significantly improved data diversity and overall model performance.

\begin{figure}
\centering
\begin{subfigure}[t]{0.20\textwidth}
\centering
\includegraphics[width=1\textwidth]{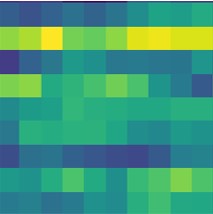}
\label{fig:mean and std of net24}
\end{subfigure}
\hspace{0.2cm}
\begin{subfigure}[t]{0.20\textwidth}
\centering
\includegraphics[width=1\textwidth]{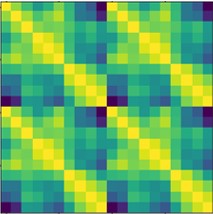}
\label{fig:mean and std of net24}
\end{subfigure}
\vspace{-.5cm}
\caption{Toeplitz covariance matrices from features samples.  This requires learning $O(n)$ parameters, in contrast to $O(n)^2$ for generic covariance matrices.}
\label{fig:mean and std of nets}
\vspace{-0.4cm}
\end{figure}

\subsection{Contributions:}
\vspace{-0.2cm}
\begin{itemize}

\item We adopted an efficient orthogonal regularization technique on convolutional layers of the few-shot classifier that enhances model generalization and intra-class feature embedding,  using the doubly block toeplitz (DBT) matrix structure.
\vspace{-0.1cm}
\item We broke down the pipeline of a few-shot learner, and based on our findings, we established three augmentations strategies namely: support augmentation, query augmentation and task augmentation that aid in minimizing overfitting.
\vspace{-0.1cm}
\item We show with compelling results that combining a DBT-based regularizer with a robust augmentation strategy improves few-shot learning performance at an average of 5\%.
\end{itemize}

\section{Related works}
\vspace{-0.12cm}
{\bf Orthogonal regularization.}
In convolutional networks, orthogonal weights have being used to stabilize layer-wise distributions and to make optimization as efficient as possible. In \cite{Bansal2018CanWG,Mishkin2016AllYN} the authors introduced orthogonal weight initialization driven by the norm preserving property of an orthogonal matrix. However, it was shown that the orthogonality and isometry property does not necessarily sustain throughout training \cite{Bansal2018CanWG} if the convolutional layers are not properly regularized. In other works, \cite{Jia2017ImprovingTO, Ozay2016OptimizationOS, Huang2018OrthogonalWN} considered Stiefel  manifold-based hard constraints of weights \cite{tagare2011notes}, but their performance reported on VGG networks \cite{Simonyan2015VeryDC} were not as promising. These aforementioned methods \cite{Jia2017ImprovingTO, Ozay2016OptimizationOS, Huang2018OrthogonalWN} are associated with hard orthogonality constraints and in most cases, they have to repeat singular value decomposition (SVD) during training which is computationally expensive on the GPUs. A recent work adopted  soft orthogonality \cite{Balestriero2018MadMA,Balestriero2018AST,Bansal2018CanWG,Xie2017AllYN}, where the Gram matrix of the weight matrix $K$ is required to be close to identity, given as $\lambda  \left\|K^TK-I  \right\|_F^2$, where $\lambda$ is the  Frobenius norm-based regularization coefficient. It's a more efficient approach than the hard orthogonality assumption \cite{Jia2017ImprovingTO, Ozay2016OptimizationOS, Huang2018OrthogonalWN, Harandi2016GeneralizedB, Xu2020LearningST} and can be viewed as a different weight decay term limiting the set of parameters close to a Stiefel manifold \cite{tagare2011notes}. Their approach constrained orthogonality among filters in one layer, leading to smaller correlations among learned features and implicitly reducing the filter redundancy. However, there are special cases where the Gram matrix cannot be close to identity which implies that matrix $K$ is overcomplete \cite{Thomas2018LearningCT}. Similarly, other works explored  orthogonal  weight  initialization \cite{Sung2018LearningTC},  mutual coherence with the isometry property \cite{Bansal2018CanWG}, penalizing  off-diagonal elements \cite{Brock2019LargeSG} towards improving kernel orthogonality.

In general, the orthogonality of $K$ alone is not sufficient to make the linear convolutional layer orthogonal among its filters.
Due to these shortcomings, we apply the improved regularization technique used in \cite{Wang2020OrthogonalCN,Le2011ICAWR}. We adopt the DBT matrix denoted as $M$ with a filter $K$, while we keep the reshaped input $x$ and output $y$ intact. The matrix multiplication; $y = Mx$ enforces the orthogonality of $M$ as shown in Figure 1 and Figure 3.

{\bf Augmentation.}
Data augmentation has become a well established technique for most  image classifiers and deep networks, as it provides an efficient strategy that significantly mitigates the models' vulnerability to overfitting. In contrast,  data augmentation still has room for expansion and adaptation in  few-shot classification or other derivatives of meta-learning in general. Existing works  \cite{Taylor2018ImprovingDL, Kang2017PatchShuffleR, Takahashi2020DataAU}, apply basic data augmentation strategies like random crops, horizontal flips and color jitter as the  staple method for most meta-learning applications. However, these aforementioned techniques have plateaued in performance with little room for significant improvement \cite{Ren2018MetaLearningFS, Ni2021DataAF}. 
Other works have added random noise to labels to alleviate overfitting \cite{Rajendran2020MetaLearningRM}, some techniques rotate all the images in a class and consider the newly rotated class as distinct from its parent class.
Recent works \cite{Dabouei2020SuperMixST, Shorten2019ASO, Ni2021DataAF, Gidaris2018DynamicFV, Qiao2018FewShotIR} are recording better performance values when augmentation strategies are injected within the meta-learning pipeline. 

In our work, we explored the benefits of including augmentation strategies along the pipeline of a DBT regularized  few-shot classifier. We identified how different augmentation approaches could affect a few-shot classifier when placed strategically along the classifier pipeline. At the core of our findings, we observed that the classifier is more sensitive to query data than support data.

{\bf Toeplitz matrix applications.}
Kimitei et al. \cite{Kimitei2011AlgorithmsFT} used toeplitz matrices  with Tikhonov regularization \cite{Natterer1984ErrorBF} as a mathematical approach to restoring blurred images. They explored their techniques  on image restoration,  enhancement, compression and  recognition. In  \cite{Hansen2004DeconvolutionAR}, the authors presented modern computational methods for treating linear deconvolution problems, they showed how to exploit the toeplitz structure to derive efficient numerical deconvolution algorithms. In compressive sensing  applications \cite{Su2014AnIT}, toeplitz-like matrices allow the entire signal to be efficiently acquired and reconstructed from relatively few measurements, compared to previous compressive sensing frameworks where a random measurement matrix is employed. 



\vspace{-0.2cm}
\begin{figure}[t!]
\begin{center}
  \includegraphics[width = 7.5cm, height= 4.5cm ]{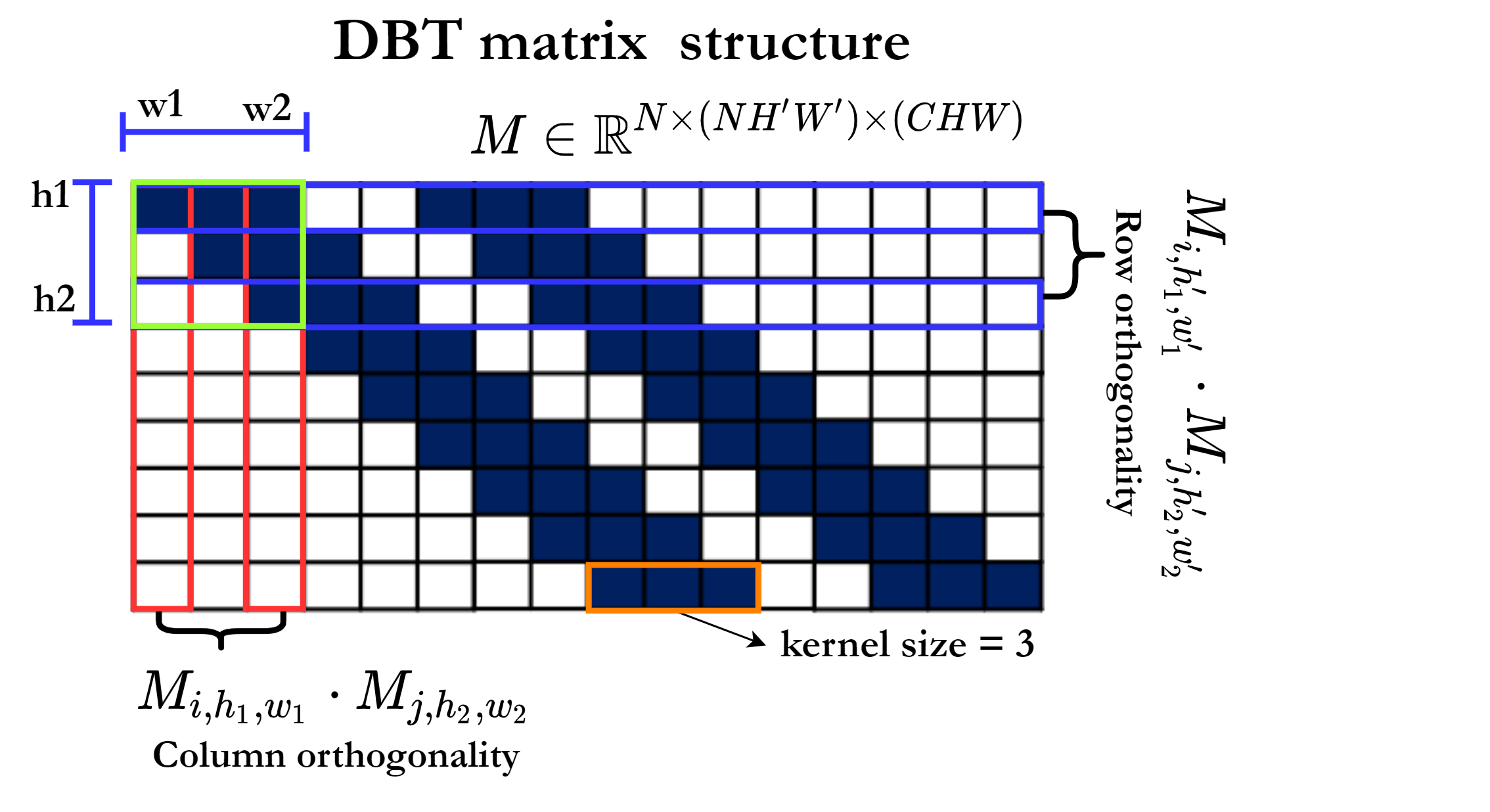}
\end{center}
\vspace{-0.5cm}
  \caption{A doubly block-Toeplitz (DBT) matrix $ M \in \mathbb{R}^{(NH'W')\times(CHW)}$ derived from the  kernel tensor  $K \in \mathbb{R}^{N \times  C\times k \times k}$.
}
\vspace{-.5cm}
\end{figure}

\section{Background}
\vspace{-0.12cm}
We consider a meta learning scenario for an  N-shot, K-way classification problem where the training and testing task datasets can be represented as $\mathcal{T} = \{ \mathcal{D}_i^{train}, \mathcal{D}_i^{test}\}_{i=1}^I$. Such a meta-training task is divided into $\mathcal{D}_{i}^{train} =\{(x_t,y_t)\}_{t=1}^{T} $ and $\mathcal{D}_{i}^{test} =\{(x_q,y_q)\}_{q=1}^{Q} $, called a meta-training set \cite{Oreshkin2018TADAMTD, Rusu2019MetaLearningWL, Ye2018LearningEA, Lee2019MetaLearningWD, Li2019FindingTF, Motiian2017FewShotAD}. The set of  $D^{train}$ and $D^{test}$ represent a small number of samples from the same distribution. 
We implement a DBT-based learner $\mathcal{B}_{dbt}(\cdot)$ to train the model for a given input feature denoted as $ y = f_{\theta}(\mathbf{x}_*)$, where (*) denotes implementations for train and test sets. We then map train and test examples into a DBT structured embedding space $\Psi_* = f_{\theta}(\mathbf{x}_{*})$. 

The objective of our model becomes:
\begin{equation}
\begin{aligned}
         & \theta = \mathcal{B}_{dbt}(\mathcal{D}_i^{train}; \phi) \\
         & \; \, = \arg \min_{\theta} \mathcal{L}^{base}(\mathcal{D}_i^{train};\theta,\phi) + \mathcal{R}(\theta),
         \label{eqa:Rterm}
    \end{aligned}
\end{equation}
\noindent
where $ \phi$ represents the parameters of the embedding model, $\mathcal{L}^{base}$ is the loss function and $\mathcal{R}$ is the regularization as described in Section 4.2.
At the end of meta-training, the performance of the model is evaluated on a set of  tasks $\mathcal{S} = \{\mathcal({D}_i^{train}, D^{test}) \}_{i=1}^I$  called the meta-testing set. The final evaluation representation over the test set is: 
\begin{equation}
\begin{aligned}
        E_{S} [L^{meta}(D^{test}; \theta , \phi)].
    \end{aligned}
    \vspace{-0.07cm}
\end{equation}
The goal of meta learning is to learn a transferable efficient embedding model $f_{\theta}$ that generalizes to new tasks. As described in section 4, we deviated from popular techniques \cite{Vinyals2016MatchingNF, Snell2017PrototypicalNF, Sung2018LearningTC, Finn2017ModelAgnosticMF} that train classifiers with convolutional blocks with some form of hard orthogonality constraint \cite{Ni2021DataAF}. Our strategy, imposes a better low displacement rank DBT-based soft orthogonality constraint on the classifier network to produce more efficient embeddings for the base learner. The final embedding model is given as:

\begin{equation}
\begin{aligned}
   \phi = \arg \min_{\phi} \mathcal{L}^{ce}(\mathcal{D}^{new};\phi),
   \end{aligned}
\end{equation}
\noindent
where $\mathcal{D}_i^{new}$ is the task from $\mathcal{T}$ and $\mathcal{L}^{ce}$ denotes the cross-entropy loss between predictions and ground truth labels.

\begin{figure*}
\begin{center}
 \hspace{1.5cm} \includegraphics[width=14.20cm, height = 4.8cm]{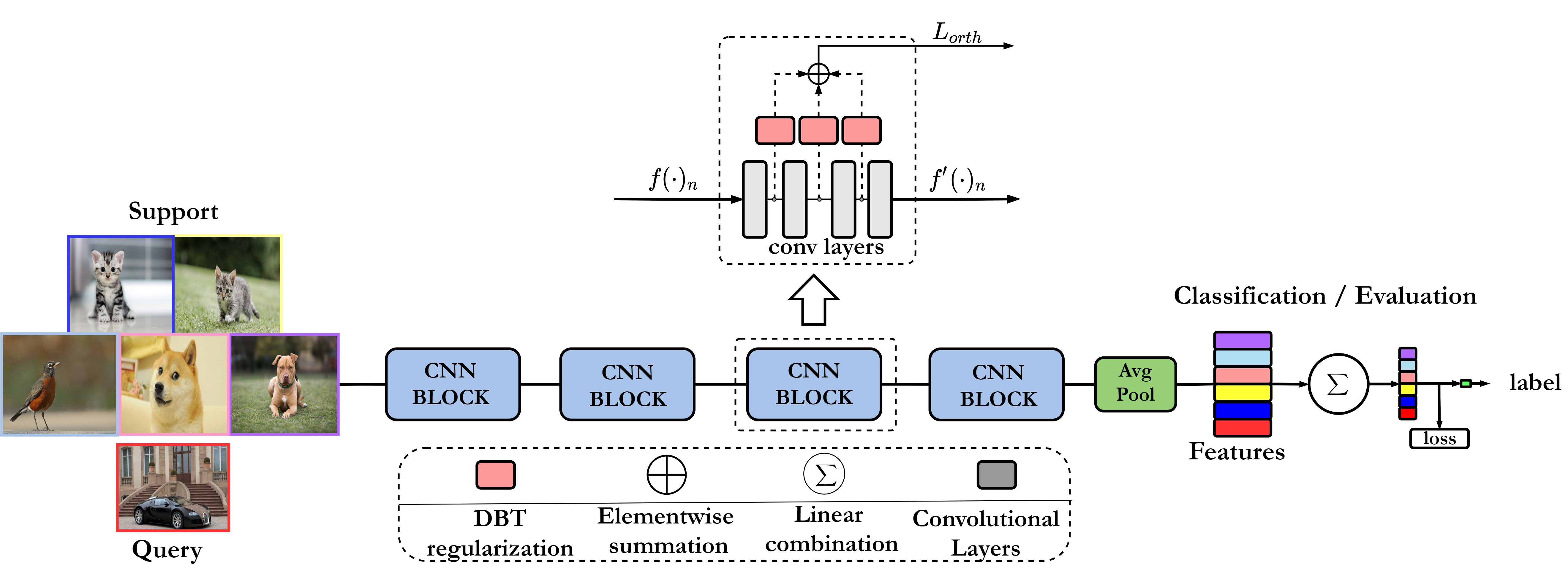}
\end{center}
\vspace{-0.4cm}
   \caption{The network depicts a DBT-regularized few shot learner. The network embeddings $B_{dbt}(\cdot)$ are regularized based on the DBT structured matrix. The dotted box of the CNN block illustrates the inner translation between convolution layer embeddings $f{(\cdot)}$ and the more efficient DBT-based embeddings denoted as $f'{(\cdot)}$, see above.  The Algorithm \ref{alg:ortho-shot} gives a logical representation of the training process.}
\label{fig:long}
\label{fig:onecol}
\vspace{-0.2cm}
\end{figure*}

\subsection{Doubly-block toeplitz (DBT) regularization}
The feature interaction between two weights vectors  $v$ and $ w $, within the layers of a few-shot classifier involves a convolution operation which can simply be represented as $ v*w = \sum \nolimits_{i=0}^{k}v(i)w(k-i)$, such that if $v$ has length $m$ and $w$ has length $n$ then $v * w$ has length $m + (n -1)$. Unfortunately, this computation involves $O(nm)$ operations which is not suitable for fast linear algebraic computations and intra-class parameter sharing which is critical for few shot learning. For such computations, if we consider a single convolution layer with input tensor $X \in \mathbb{R}^{ C \times H \times W}$ and kernel $K \in \mathbb{R}^{N \times  C\times k \times k}$,  the convolution’s output tensor is expressed as $Y = Conv(K, X)$, where $Y \in \mathbb{R}^{N \times  H' \times W'} $, we replaced the convolution operator $(*)$ with $Conv(.)$ for simplicity. $N$, $H$, $W$ and $C$ are the number of kernels, height, width and channel of the input tensor, respectively. While $k$ represents the kernel size and $H'$, $W'$ are the height and width for the output tensor, respectively.

Inline with our goal to improve the computational complexity and enhance better feature representation, we  adapted  a DBT matrix construction by utilizing the linear property of the convolution operation. The convolution expression; Conv(K, X) is converted into a faster DBT matrix-vector representation given as: 
\begin{equation}
\begin{aligned}
            Y = Conv(K,X) \Leftrightarrow y = Mx. 
            \label{eqa:dbt}
    \end{aligned}
\end{equation}
\noindent
This simple rearrangement establishes the foundation for adapting the DBT regularizer in our few-shot classifier network. Where $M$ is the DBT matrix, $x$ and $y$ represent flattened input and output tensors, respectively. $M$ is structured and is of rank $r<<min(m,n)$ \cite{Thomas2018LearningCT}, this representation minimizes the storage requirements to $(mr+nr)$ parameters and accelerates the matrix-vector multiplication time to $O(mr +nr)$. Section 1: Figure 1 in the supplementary material shows the hierarchy for storage cost and operation count for matrix-vector multiplications. This DBT formulation stabilizes the spectrum of the newly derived DBT-based matrix $M$. In section 1.1 and 1.4 of the supplementary material, we reflect the overall benefit of the DBT model.

\begin{algorithm}
\small
\caption{Ortho-shot algorithm}\label{alg:ortho-shot}
\begin{algorithmic}[1]
\Require {}$ \mathcal{D} \gets \{\mathcal{X},\mathcal{Y}\}_{i=1}^{N}$

\Procedure{Orthogonal-regularizer}{}
\State $ I_{ro} \gets t(I)$ \Comment toeplitz matrix
\State $ M \gets Conv(K,K)$ 
\State $ y \gets  \lambda(\left\| M - I_{ro} \right\|) $
\State $ {\psi}(\cdot) \gets Mx $  \Comment DBT based output
\EndProcedure
\Require{} $ \mathcal{D}^{train}, \mathcal{D}^{test} \gets \{\mathcal{X}^t, \mathcal{Y}^t \, ; \,\mathcal{X}^q, \mathcal{Y}^q\}_{i=1}^{K}$

\Procedure{Few Shot Learning}{}
\If{Train}
\For{$i=1$}, T
\State $ B_{dbt}\gets  S_{\psi}  $ \Comment DBT model
\State $\mathcal{L}^{dce} + \mathcal{L}^{orth} \gets loss(B_{dbt}(\mathcal{X}^t), \mathcal{Y}^{t})$
\State $ \mathcal{L}^{total}_{dbt} \gets \mathcal{L}^{dce} + \lambda \mathcal{L}^{orth} $
\State \Comment (orthogonal \,\, regularization) 
\EndFor 
\EndIf
\If{Test}
\For{$i=1$}, Q
\State $\mathcal{L}^{test}  \gets loss(B_{dbt}(\mathcal{X}^q), Y^{q})$
\EndFor 
\EndIf
\EndProcedure
\end{algorithmic}
\vspace{-0.1cm}
\end{algorithm}
\vspace{-0.5cm}

\section{The proposed method}
We present an  efficient  low displacement  rank  (LDR)  regularization  strategy  termed Ortho-Shot that imposes orthogonal regularization  on  the  convolutional  layers  of  a  few-shot  classifier  which  is  based  on  the  doubly-block  toeplitz  (DBT) matrix  structure \cite{Wang2020OrthogonalCN, Huang2018OrthogonalWN}.  Our technique, as reflected in section 4.1 deviates from popular methods that train classifiers with convolutional  blocks  with  some  form  of  hard  orthogonality  constraint. We also adapted a set of augmentation strategies based on the support, query and task datasets to boost overall model performance. In general, our approach  enhances model generalization, intra-class feature embeddings and also minimizes overfitting for a few-shot classifier. To further describe our approach, we consider a single convolutional layer case. We extract feature embeddings  $X \in \mathbb{R}^{C \times H \times W} $ from the intermediate convolutional layers of the few-shot classifier and then flatten it to a vector  $x \in \mathbb{R}^{1 \times (H \times W)} $. The weight tensor; $K$ of our model is also converted to a doubly block-Toeplitz (DBT) matrix $ M \in \mathbb{R}^{(NH'W')\times(CHW)}$ derived from kernel tensor  $K \in \mathbb{R}^{N \times  C\times k \times k}$ as shown in Figure 3. With the aforementioned matrix structure, we are able to apply a better orthogonality constraint as described by the Lemma in \ref{l1}. In Figure 4, we show a fully regularized setup for a single CNN block. The network embeddings $B_{dbt}(\cdot)$ are regularized based on the DBT structure and the entire losses from each respective layer is summed up to $L_{orth}$. We show promising results for our technique  as described by the CAM plots in Figure 5.

\subsection{Convolutional orthogonality}
A DBT kernel matrix $M$ can be applied on both a rectangular or square case, where kernel   $ M \in \mathbb{R}^{(NH'W')\times(CHW)}$
dimensions can be rectangular $(NH'W') \leq(CHW)$ or square, $(NH'W')  > (CHW)$. In the rectangular case, the uniform spectrum applies row orthogonal convolution while the square case requires column orthogonal convolution. In theory, the DBT kernel $M$ is highly structured  and sparse \cite{Le2011ICAWR} as  a result, an equivalent representation is required to regularize the spectrum of $M$ to be uniform \cite{Wang2020OrthogonalCN,Huang2018OrthogonalWN}. We give the cases for both row and column orthogonality and we also propose an equivalent representation in this section.

{\bf Row orthogonality case.} 
The row of matrix $M$ corresponds to a filter at a particular spatial location flattened to a vector, denoted as $M_{i,h'w'} \in  \mathbb{R}^{CHW}$.  The row orthogonality condition is given as:
\vspace{-.3cm}
\begin{equation}
  \begin{aligned}
\langle {M_{i, h'_{1}, w'_1} \cdot  M_{jh'_{2}, w'_2}}\rangle = 
    \begin{cases}
      1, &  i_{h'_{1}, w'{_1}} =  j_{h'_{2}, w'{_2}} \\
      0, &  otherwise.
    \end{cases}  
    \end{aligned} 
\end{equation}
This results to an equivalent of Equation \ref{eqa:dbt} as the following self-convolution: 
\begin{equation}
\begin{aligned}
    Y = Conv(K, K, padding = P, stride = S) = I_{r0},
    \label{eqa:lorth}
    \end{aligned}
    \end{equation}
where  $ I_{r0} \in \mathbb{R}^{ N \times N \times (2P/S+1) \times (2P/S+1)}$ is a tensor with an identity matrix at the centre and zeros entries elsewhere.

\begin{figure*}
\begin{center}
  \includegraphics[width=17.9cm, height = 4.6cm ]{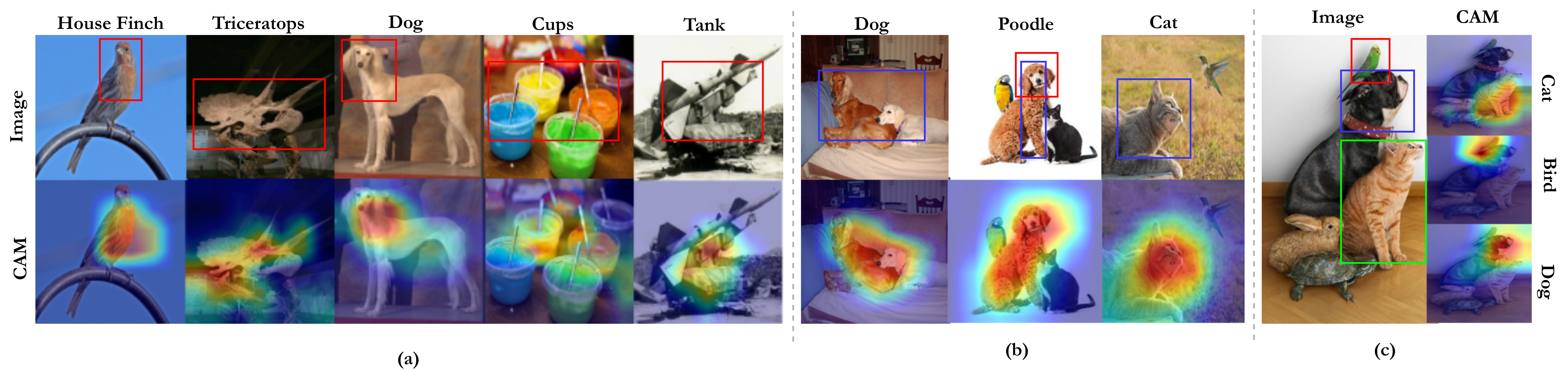}
\end{center}
\vspace{-.5cm}
   \caption{Illustration of CAM plots. (a) The second row represents CAM plots for single classes. The red squares highlight regions of interest clearly highlighted by the model.(b) Shows more complex scenarios where multi classes are involved. Classes are clearly separated from non-classes of interest. In (c), all the objects separated by  bounding boxes are clearly localised as indicated by the CAM plot.}
\label{fig:long}
\label{fig:onecol}
\vspace{-.4cm}
\end{figure*}

{\bf Column orthogonality case.}
If  $X_{i,hw} \in \mathbb{R}^{C \times H \times W} $ denotes an input tensor, which has all zero except an entry at the $i^{th}$ input channel at spatial location $(h, w)$. Then we can denote the flattened vector as  $x_{ihw} \in \mathbb{R}^{C \times H \times W} $ derived from  $X_{i,hw}$. A column vector  $M_{i,hw}$  of $M$  is obtained by multiplying $M$ and column vector $x_{i,hw}$.
Similar to the row orthogonality, 
\begin{equation}
\begin{aligned}
    Y = Conv(K^T, K^T, padding = k-1, stride = 1) = I_{c0},
    \end{aligned}
    \end{equation}
where $K^T$ is the input-output transposed $K$, i.e., $ K^T \in \mathbb{R}^{ C \times  N\times k \times k}, I_{c0} \in \mathbb{R}^{ C \times  C\times(2k-1) \times (2k-1) }  $ has all zeros except for the center $C \times C$ entries as an identity matrix. Figure 2 illustrates the DBT matrix $M$ structure of our model.

\subsection{Row-column orthogonality equivalence}
To develop an equivalent representation for row and column orthogonality, we build on the equation described by {\bf lemma 1}, which states that the minimizing of the column orthogonality and row orthogonality costs are equivalent \cite{Le2011ICAWR} due to the property of the Frobenius norm. \newline
\vspace{-0.2cm}
\newline
\label{l1}{\bf Lemma 1:} The row orthogonality cost $\lambda  \left\|KK^T-I_{r0}\right\|_F^2  $  is equivalent to the column orthogonality cost   $\lambda  \left\|K^TK-I_{c0}\right\|_F^2 + U$ where $U$ is a constant. This implies that convolution orthogonality independent of the shape of $M$ (square or rectangular) can be regularized, given as:
\vspace{-0.2cm}
\begin{equation}
\begin{aligned}
           L_{orth} = \lambda \left\|K^TK-I_{r0}\right\|_F^2,
    \end{aligned}
\end{equation}
\noindent
where $L_{orth}$ is the DBT-based orthogonal regularization term that depends only on Equation \ref{eqa:lorth} and replaces the $\mathcal{R(\cdot)}$ term in Equation \ref{eqa:Rterm}.

\section{Experimental setup and analysis}
Our experiments were conducted on the miniImagNet, CIFAR-FS, Stanford Dogs and Stanford Cars datasets, respectively. We used the R2-D2 base leaner \cite{Bertinetto2019MetalearningWD}, the "ResNet-12" and "64-64-64-64" backbone for different few-shot learning modes used in our work. Data augmentation strategies were also analysed to determine the best combination for a DBT-regularized model. The complete details for of the entire setup is expressed in section 1.2 of the supplementary material.

\subsection{Data augmentation strategy}
Motivated by the impact  of applying a diverse augmentation strategy on meta-learners, we established three unique augmentation approaches; support, query and task augmentation that contribute to the overall classifier performance, aimed at minimising overfitting. Our empirical analysis confirm that support augmentation increases the number of fine tuning data while the query data improves evaluation performance while training the classifier. Similarly, task augmentation is used to increase the number of classes per task while training. We adapted a couple of augmentation techniques such as CutMix \cite{Yun2019CutMixRS}, where image patches  are  cut  and  pasted  among  training  images  and  the  ground  truth  labels  are  also  mixed proportionally  within  the  area  of  the  patches. Mixup \cite{Seo2021SelfAugmentationGD}, a technique that generates convex combinations of pairs of examples and their labels, which proved to be effective for support and query augmentation strategies. As well as Self-Mix \cite{Zhang2018mixupBE}  in which an image is substituted into other regions in the same image. This dropout effect improves few-shot learning generalization overall. In addition, we implemented standard data augmentation techniques by randomly erasing patches from the images (Random Erase), horizontally flipping the images (Horizontal Flip), rotating the images at different specified angles (Rotation)  and Color Jitter, where we randomly change the brightness, contrast and saturation of the images. To boost the performance of our augmentation strategy, we combine different augmentation techniques using the MaxUp augmentation approach proposed in \cite{Gong2020MaxUpAS}. The rationale behind MaxUp augmentation is to minimize training loss by performing parameter updates on the task that maximizes loss in a min-max optimization manner, the MaxUp expression is  given as:
\begin{equation}
\begin{aligned}
       \min_{\theta}  E_{\mathcal{T}} [ \max_{M \in \mathcal{S}}  \mathcal{L}( \mathcal{B}_{\theta'}, M(\mathcal{T}^q))],
    \end{aligned}
\end{equation}
\noindent 
where $\theta'$ represents the model parameters, $\mathcal{B}$ is the base model, $\mathcal{L}$ is the loss function and $\mathcal{T}$ is a task for both support and query data;  $\mathcal{T}^s$ and $\mathcal{T}^q$, respectively. 


\subsection{Augmentation performance}
In this section, we investigate the performance of a few-shot classifier for different  augmentation strategies. We investigated three test cases that check the training performance when data is sampled from the support, query and task data, respectively. Our approach is similar to techniques adapted by \cite{Ni2021DataAF, Kye2020TransductiveFL, Seo2021SelfAugmentationGD} that examine the impact of augmentation on a diverse set of data combinations.

{\bf Case 1:} We trained the model at an equal number of support and query data as indicated in Table 1, so as to establish a baseline performance of the model. We use this strategy to  compare the impact of any of the data pools (support or query) when any of the augmented pairs is reduced.

{\bf Case 2:} We initiated training of the classifier by randomly sampling from 5 and 10 unique samples per class of the  support data while using the entire query data pool. Using this approach, we reduced the influence of support data in order to examine the impact of the diverse pool of query data on the classifier. Our findings reflected in Table 1 show accuracy values at $\pm $ 2\%. This is a clear indication that augmentation of query data plays a more significant role in the overall model performance. In contrast, we reduced the number of query data while maintaining the initially set cap for support data and recorded a decline in accuracy.

{\bf Case 3:} To evaluate the impact of task augmentation, we used the CIFAR-FS data to initially allocate 10 distinct 5-way classification tasks (252  combinations) before training, while the support and query datasets are maintained equally at 500, respectively. We observed a decline in performance. However, as we increased the amount of task data, significant improvement is recorded, which confirms that task augmentation is crucial in few-shot learning.

In summary, we broke down the few-shot learning process to determine the influence of support, query and task augmentation, respectively. Our findings confirm that our baseline learner is most sensitive to query data \cite{Ni2021DataAF}. In addition, task augmentation provided significant value (about 2\%) that cannot be overlooked by the classifier.

\begin{table}
\begin{center}
\vspace{-.1cm}
\caption{Few-shot classification accuracy (\%) using R2-D2 base leaner  with a  ResNet-12  backbone on CIFAR-FS dataset. Support, Query and Task columns represent the number of samples per class for support, query data and the total  number of tasks available.}
\begin{adjustbox}{height=1.55cm, width=7.8cm}
\begin{tabular}{lcccc} 
\hline
\textbf{ Support} &  \textbf{Query} &  \textbf{Task} &  \textbf{1-shot} &  \textbf{5-shot} \\
\hline
500  & 500 & full & 71.41$\pm$ 0.21 & 86.01$\pm$ 0.08\\ 
100  & 500 & full & 70.11$\pm$ 0.01 & 83.00$\pm$ 0.03\\ 
10   & 500 & full & 70.72$\pm$ 0.01 & 81.41$\pm$ 0.32\\ 
\hline
500  & 300 & full & 69.41$\pm$ 0.11 & 72.41$\pm$ 0.08\\ 
500  & 100 & full & 59.00$\pm$ 0.21 & 70.41$\pm$ 0.08\\ 
\hline
5 (random)  & 500 & full & 61.01$\pm$ 0.11 & 80.41$\pm$ 0.08\\ 
10 (random) & 500 & full & 63.01$\pm$ 0.30 & 81.24$\pm$ 0.02\\ 
\hline
\end{tabular}
\end{adjustbox}
\end{center}\bf
\vspace{-.4cm}
\end{table}

\begin{figure}
\centering
\begin{subfigure}[t]{0.235\textwidth}
\centering
\includegraphics[width=1\textwidth]{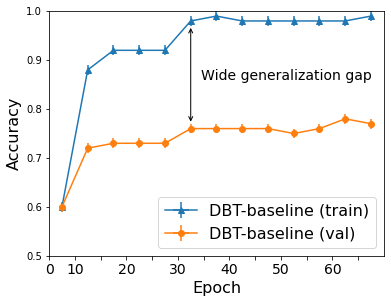}
\label{fig:mean and std of net24}
\end{subfigure}
\begin{subfigure}[t]{0.235\textwidth}
\centering
\includegraphics[width=1\textwidth]{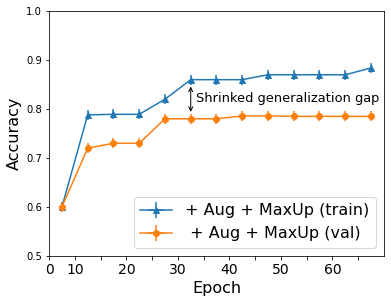}
\label{fig:mean and std of net24}
\end{subfigure}
\vspace{-0.3cm}

\begin{subfigure}[t]{0.234\textwidth}
\centering
\includegraphics[width=1\textwidth]{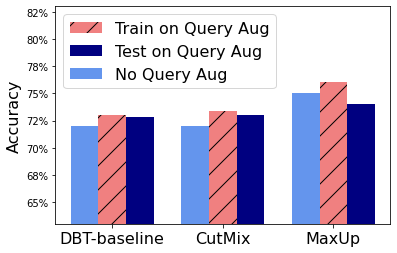}
\label{fig:mean and std of net24}
\end{subfigure}
\begin{subfigure}[t]{0.226\textwidth}
\centering
\includegraphics[width=1\textwidth]{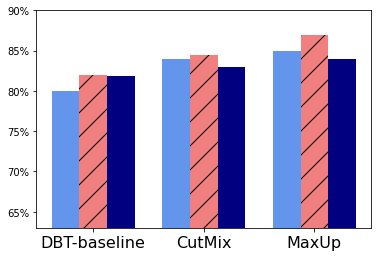}
\label{fig:mean and std of net24}
\end{subfigure}
\vspace{-0.5cm}
\caption{Accuracy results for training and validation on R2-D2 base-learner \cite{Bertinetto2019MetalearningWD} with a DBT-regularized ResNet-12 backbone on the CIFAR-FS dataset. (Top Left) Baseline model and (Top right) Augmentation "Aug" and MaxUp. The MaxUp augmentation strategy narrows down the generalization gap and reduces overfitting. (Bottom left) 1-shot classification and (Bottom Right) 5-shot classification for Query data augmentation.}
\label{fig:mean and std of nets}
\vspace{-0.3cm}
\end{figure}

\begin{figure}
\centering
\begin{subfigure}[t]{0.234\textwidth}
\centering
\includegraphics[width=1\textwidth]{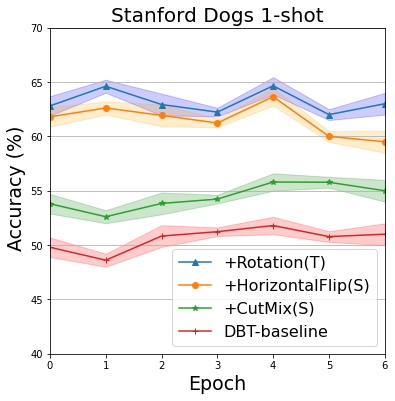}
\label{fig:mean and std of net24}
\end{subfigure}
\begin{subfigure}[t]{0.220\textwidth}
\centering
\includegraphics[width=1\textwidth]{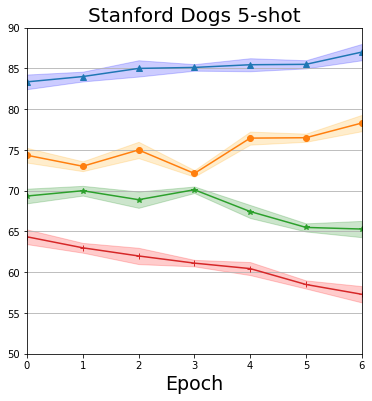}
\label{fig:mean and std of net24}
\end{subfigure}
\vspace{-0.5cm}
\caption{ Accuracy plots for different datasets compared to the baseline model. Augmentations techniques were applied on Task "T" and Support "S" datasets. Overall, accuracy for 5-shot is maintained at 85-88\% while for 1-shot, a range of 65-68\% is recorded.}
\label{fig:mean and std of nets}
\end{figure}

\begin{table}
\begin{center}
\caption{Few-shot classification accuracy (\%) using R2-D2 base leaner with a  ResNet-12  backbone on the  CIFAR-FS dataset. Support(S), Query(Q) and Task(T) data are used on different augmentation strategies.}
\begin{adjustbox}{height=2.6cm, width=6.3cm}
\begin{tabular}{lccc} 
\hline
\textbf{Augmentation} & \textbf{1-shot} &  \textbf{5-shot}\\
\hline
CutMix(Q)     &  76.01$\pm$ 0.21 & 87.14$\pm$ 0.08\\ 
+ CutMix(S)   &  75.11$\pm$ 0.31 & 85.30$\pm$ 0.14\\ 
+ Horizontal Flip(S)  &  76.32$\pm$ 0.11 & 87.01$\pm$ 0.23\\ 
+ Rotation(T)  &  75.33$\pm$ 0.25 & \textbf{87.68$\pm$ 0.03}\\ 
\hline
SelfMix(Q)    &  76.04$\pm$ 0.21  & 86.81$\pm$ 0.08\\ 
+ CutMix(S)   &   76.19$\pm$ 0.29 & 86.35$\pm$ 0.16\\ 
+ Horizontal Flip(S)  &  75.27$\pm$ 0.32 & 86.88$\pm$ 0.03\\ 
+ Rotation(T)  &  75.61$\pm$ 0.22 & \textbf{87.40$\pm$ 0.18}\\ 
\hline
MixUp(Q)      &  72.14$\pm$ 0.01  & 82.81$\pm$ 0.08\\ 
+ CutMix(S)   &   71.03$\pm$ 0.29 & 85.15$\pm$ 0.11\\ 
+ Horizontal Flip(S)  &  72.27$\pm$ 0.10 & 83.08$\pm$ 0.01\\ 
+ Rotation(T) &  74.10$\pm$ 0.11  & 85.10$\pm$ 0.22\\
\hline
\end{tabular}
\end{adjustbox}
\end{center}\bf
\vspace{-0.7cm}
\end{table}

\begin{table*}
\begin{center}
\caption{
{Comparison to prior work on miniImageNet and CIFAR-FS}. Few-shot classification accuracy (\%) using R2-D2 base leaner a "ResNet-12" and "64-64-64-64" backbone on CIFAR-FS and miniImageNet datasets, respectively. We applied Rotation (R) to the CutMix and Horizontal Flip (HF) to the SelfMix augmentation modes. "Q" denotes query data, "S" represents support data and "M" dentotes MaxUp. 
}
\vspace{-0.15cm}
\begin{adjustbox}{height=2.29cm, width=10.80cm}
\begin{tabular}{lccccc} 
\hline
& \multirow{2}{*} & \multicolumn{2}{c}{\textbf{CIFAR-FS 5-way}} &  \multicolumn{2}{c}{\textbf{miniImageNet 5-way}}\\
\cline{2-6}\textbf{DBT-model} &  \textbf{Backbone} &  \textbf{1-shot} &  \textbf{5-shot} &  \textbf{1-shot}  &  \textbf{5-shot} \\
\hline
Basline(No Aug) & ResNet-12 & 70.26 $\pm$ 0.61 &  83.12 $\pm$ 0.53 & 55.03 $\pm$ 0.40  & 74.06 $\pm$ 0.24 \\ 
CutMix(Q) & ResNet-12 & 71.46 $\pm$ 0.24 &  84.32 $\pm$ 0.73 & 57.36 $\pm$ 0.24  & 74.46 $\pm$ 0.11\\ 
CutMix(Q) + M & ResNet-12  & \textbf{72.00 $\pm$ 0.01}& \textbf{86.20 $\pm$ 0.61}  &  \textbf{58.13 $\pm$ 0.25} & \textbf{75.69 $\pm$ 0.74}\\
\hline
SelfMix(S) + R & ResNet-12  & 62.56 $\pm$ 0.54 &  79.82 $\pm$ 0.33 & 50.38 $\pm$ 0.63 & 71.44 $\pm$ 0.08  \\ 
SelfMix(S) + M &  ResNet-12  & \textbf{63.51 $\pm$ 0.78}& \textbf{80.20 $\pm$ 0.66}  &  \textbf{57.31 $\pm$ 0.89} &\textbf{72.69 $\pm$ 0.70}\\
\hline
CutMix(S) + HF & 64-64-64-64 & 60.56 $\pm$ 0.29 &  85.32 $\pm$ 0.73 & 62.26 $\pm$ 0.63 & 79.28 $\pm$ 0.63  \\ 
CutMix(S) + M &  64-64-64-64 & \textbf{63.42 $\pm$ 0.17} & \textbf{86.33 $\pm$ 0.66}  &  \textbf{63.31 $\pm$ 0.89} & \textbf{80.69 $\pm$ 0.54}\\
\hline
SelfMix(Q) + HF & 64-64-64-64 & {75.56 $\pm$ 0.84} &  {84.32 $\pm$ 0.73 }&  {66.31 $\pm$ 0.89} & {82.69 $\pm$ 0.74}  \\ 
SelfMix(Q) + M &  64-64-64-64 & \textbf{76.42 $\pm$ 0.38} & \textbf{86.10 $\pm$ 0.36}  &  \textbf{67.39 $\pm$ 0.34} & \textbf{83.44 $\pm$ 0.24}\\
\hline
\end{tabular}
\end{adjustbox}
\end{center}\bf
\vspace{-0.5cm}
\end{table*}

\begin{table*}
\begin{center}
\caption{Experimental results that compare prior work on the Stanford Dogs, Stanford Cars and  CIFAR-FS dataset. Average few-shot classification accuracy with 95 \% confidence intervals. The second column shows which kind of embedding is employed, we used a 4-layer convolutional network with their respective filters in each layer.}
\vspace{-0.1cm}
\begin{adjustbox}{height=3.1cm, width=15.30cm}
\begin{tabular}{lcccccc} 
\hline
 & \multicolumn{2}{c}{\textbf{Stanford Dogs 5-way}} &  \multicolumn{2}{c}{\textbf{Stanford Cars 5-way}} & \multicolumn{2}{c}{\textbf{CIFAR-FS 5-way}} \\
\cline{2-7}\textbf{ Model}  &  \textbf{1-shot} &  \textbf{5-shot} &  \textbf{1-shot}  &  \textbf{5-shot} &  \textbf{1-shot} &  \textbf{5-shot}\\
\hline
Matching Networks \cite{Vinyals2016MatchingNF} & 35.80 $\pm$ 0.99 & 47.50 $\pm$ 1.03 & 34.80 $\pm$ 0.98 & 44.70 $\pm$ 1.03 &61.16 $\pm$ 0.89 &72.86 $\pm$ 0.70 \\ 
MAML \cite{Finn2017ModelAgnosticMF}  & 44.81 $\pm$ 0.34 & 58.68 $\pm$ 0.31 & 47.22 $\pm$ 0.39 & 61.21 $\pm$ 0.28 & 55.92 $\pm$ 0.95 & 72.09 $\pm$ 0.76\\ 
Relation Nets \cite{Sung2018LearningTC} & 43.33 $\pm$ 0.42 & 55.23 $\pm$ 0.41 & 47.67 $\pm$ 0.47 & 60.59 $\pm$ 0.40 & 62.45 $\pm$ 0.98 & 76.11 $\pm$ 0.69\\
Prototypical Networks \cite{Snell2017PrototypicalNF}  & 37.59 $\pm$ 1.00 & 48.19 $\pm$ 1.03 & 40.90 $\pm$ 1.01 & 52.93 $\pm$ 1.03 & 51.31 $\pm$ 0.91 & 70.77 $\pm$ 0.69\\ 
DN4 \cite{Li2019RevisitingLD}  & 45.41 $\pm$ 0.76 & 63.51 $\pm$ 0.62 & 59.84 $\pm$ 0.80 & 88.65 $\pm$ 0.44 & 52.79 $\pm$ 0.86 & 81.45 $\pm$ 0.70\\
PABN \cite{Huang2019LowRankPA}  & 45.65 $\pm$ 0.71 & 61.24 $\pm$ 0.62 & 54.44 $\pm$ 0.71 & 67.36 $\pm$ 0.61 & 63.56 $\pm$ 0.79 & 75.35 $\pm$ 0.58\\
MATANet \cite{Chen2020MultiscaleAT} & 55.63 $\pm$ 0.88 & 70.29 $\pm$ 0.62 & 73.15 $\pm$ 0.88 & 91.89 $\pm$ 0.45 & 67.33 $\pm$ 0.84 & 83.92 $\pm$ 0.63\\
GNN \cite{Satorras2018FewShotLW}  & 46.38 $\pm$ 0.78 & 62.27 $\pm$ 0.95 & 55.85 $\pm$ 0.97 & 71.25 $\pm$ 0.89 & 51.83 $\pm$ 0.48 & 63.69 $\pm$ 0.94\\
Rfs \cite{Tian2020RethinkingFI} & 55.64 $\pm$ 0.28 & 62.02 $\pm$ 0.63 & 79.64 $\pm$ 0.44  &   69.74 $\pm$ 0.72 & 83.41 $\pm$ 0.55 & 83.50 $\pm$ 0.11\\
Rfs-distill \cite{Tian2020RethinkingFI} & 56.01 $\pm$ 0.48 & 64.82 $\pm$ 0.60 & \textbf{82.14 $\pm$ 0.43}   &   71.52 $\pm$ 0.69 & \textbf{86.03 $\pm$ 0.49} & 84.10 $\pm$ 0.28\\
\hline
DBT-baseline & {56.06 $\pm$ 0.03}  &  { 71.00 $\pm$ 0.25} & {73.49 $\pm$ 0.01} & {92.02 $\pm$ 0.33} & {74.41 $\pm$ 0.50} & {84.21 $\pm$ 0.65}\\

+ CutMix(Q) + R  & {56.36 $\pm$ 0.64}  &  { 71.39 $\pm$ 0.04} & {73.69 $\pm$ 0.51} & {93.00 $\pm$ 0.15} & {74.81 $\pm$ 0.37} & {86.01 $\pm$ 0.67}\\

+ SelfMix(Q) + HF  & {56.86 $\pm$ 0.64}  &  { 72.19 $\pm$ 0.78} & {74.21 $\pm$ 0.01} & {93.30 $\pm$ 0.35} & {75.01 $\pm$ 0.15} & {87.01 $\pm$ 0.74}\\

+ MaxUp & \textbf{57.06 $\pm$ 0.63}  &  \textbf{ 73.15 $\pm$ 0.22} & {75.34 $\pm$ 0.41} & \textbf{94.38 $\pm$ 0.25} & {76.41 $\pm$ 0.25} & \textbf{87.68 $\pm$ 0.24}\\
\hline
\end{tabular}
\end{adjustbox}
\end{center}\bf
\vspace{-0.40cm}
\end{table*}

\subsection{Augmentation modes}
\vspace{-0.1cm}
This section builds on the findings of section 5.1, where we established three core data augmentation cases; support, query and task data augmentation. Similar to \cite{Ni2021DataAF, Gong2020MaxUpAS, Dabouei2020SuperMixST}, we used the CutMix, SelfMix, MixUp, Random Crop and Horizontal Flip augmentation methods on the support, query and task datasets, respectively. We identified the best augmentation combinations that suit a few-shot learner and with our findings, we picked the best  strategy to determine which mode of augmentation suits a DBT-regularized few-shot learner. To start with, we used the R2-D2 base learner \cite{Bertinetto2019MetalearningWD} and the CIFAR-FS database to evaluate the augmentation performance on support, query and task augmentations as shown in Table 1. Our findings show that the pair of CutMix and SelfMix augmentation produces the best results with over 2.5\% in accuracy improvement \cite{Ni2021DataAF}. Other approaches lag behind in performance at about $\pm $ 3\% for both 1-shot and 5-shot cases.  Secondly, since the CutMix and SelfMix methods stand out as the best augmentation approach for our setup, we used them as bases to combine augmentations on the three data cases; support, query and task, respectively as shown in Table 2. Model performance significantly improved with the best case occurring when CutMix (query) is combined with SelfMix (support).
\begin{figure}
\centering
\begin{subfigure}[t]{0.220\textwidth}
\centering
\includegraphics[width=1\textwidth]{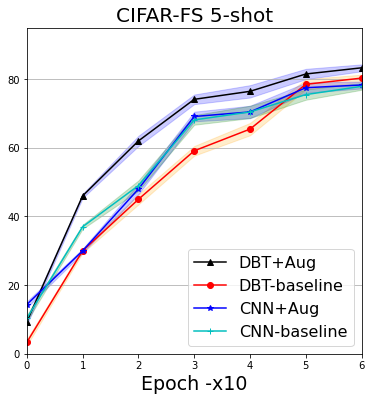}
\label{fig:mean and std of net24}
\end{subfigure}
\begin{subfigure}[t]{0.220\textwidth}
\centering
\includegraphics[width=1\textwidth]{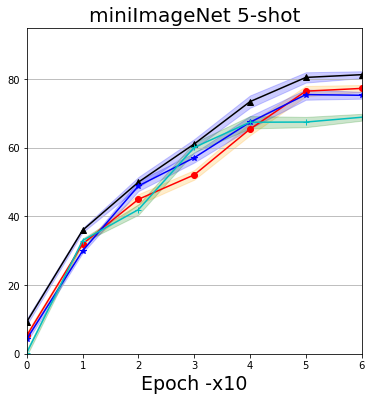}
\label{fig:mean and std of net24}
\end{subfigure}
\begin{subfigure}[t]{0.220\textwidth}
\centering
\includegraphics[width=1\textwidth]{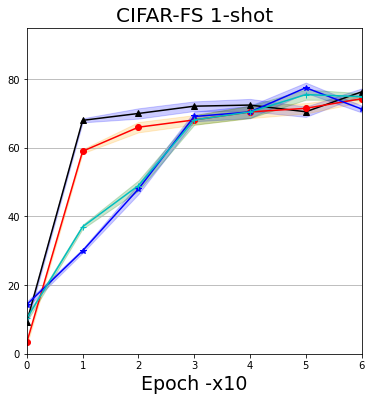}
\label{fig:mean and std of net24}
\end{subfigure}
\begin{subfigure}[t]{0.220\textwidth}
\centering
\includegraphics[width=1\textwidth]{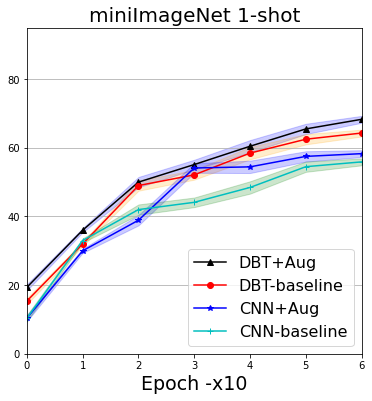}
\label{fig:mean and std of net24}
\end{subfigure}
\vspace{-0.4cm}
\caption{ Model accuracy plots for CIFAR-FS and miniImageNet datasets on CNN and DBT model baselines with augmentation "Aug" and without Augmentation for 5-shot and 1-shot cases.
}
\label{fig:mean and std of nets}
\vspace{-0.35cm}
\end{figure}
\vspace{0.3cm}
\subsection{DBT-regularization with data augmentation}
As discussed in section 1, DBT-based regularization improves model generalization and intra-class feature expressiveness. Data augmentation on the other hand creates sufficient data diversity which helps to mitigate overfitting. In this section, we highlight the collective benefits of combing a DBT-based regularizer with augmentation strategies for few-shot learning, using different datasets. 

\noindent{\bf Accuracy results with different datasets:}
In this section, we setup a testing scheme where we evaluate our method over four runs which is quite similar to techniques applied in \cite{Tian2020RethinkingFI}. We computed the mean accuracy as the accuracy for every run, the experiments are conducted on the Stanford Dogs, Stanford Cars, miniImageNet and CIFAR-FS datasets, respectively as shown in Table 4 and Table 3. Our accuracy results for 5-shot were  maintained at 80-88\% while for 1-shot at a range of 65-68\%  as shown in Figure 7 and Figure 8.  Our baseline integrated with the DBT-based regularizer "DBT-baseline" model performs at about 2\% better than state-of-the art without data augmentation. Applying the CutMix and SelfMix augmentation on the query "Q" and support "S" datasets, show significant improvement. Rotation "R" and Horizontal Flip "HF"  are integrated into the CutMix and SelfMix data augmentation modes,  respectively as indicated in Table 3.

\noindent{\bf Improvement with MaxUp augmentation:}
In this section, we evaluate the performance of our model with the Max-Up approach for both 1-shot and 5-shot classification. We use a similar experimentation setting described in \cite{Ni2021DataAF} at different augmentation pool sizes. Figure 6 and Table 4 depict the impact of Max-Up with the augmentation strategies; denote generally as  "Aug" for both train and validation data. We also show results  for the baseline model without augmentation (DBT-baseline), with CutMix  and MaxUp augmentation for  different Query data schemes. We observe from Figure 6 (Top right) that the generalization gap shrinks considerably  and by implication, overfitting is minimized when  the MaxUp strategy is adapted. MaxUp also adds the extra boost with an average of about 2.3\% in performance.

\noindent{\bf Comparison with different methods:}
We compared our results against different methods  \cite{Satorras2018FewShotLW, Chen2020MultiscaleAT, Huang2019LowRankPA, Li2019RevisitingLD} as shown in Table 4. We observed that \cite{Tian2020RethinkingFI} is closest to ours but we outperform their approach significantly for the 5-shot cases by over 6\% on the average. We recorded a better performance than GNN \cite{Satorras2018FewShotLW} and MATANet \cite{Chen2020ANM} using both the  5-way 1-shot and 5-way 5-shot few-shot learning settings, we saw an improvement of about 3.3\%  4.2\% and 3.16\% on Stanford Dogs,  Stan-ford  Cars,  and  CIFAR-FS,  respectively for 5-way 1-shot task while for  the  5-way5-shot task,  our method achieved about 4.7\%,  2.1\%,  and 4.9\%  overall. Clearly, the MaxUp boost is significant in almost all cases.

\vspace{-0.1cm}
 \section{Conclusion}
We proposed a structured doubly block-toeplitz (DBT) matrix based model that imposes orthogonal  regularization  on the filters of the convolutional layers termed  Ortho-Shot. Our approach was aimed at  maintaining  the  stability  of  activations,  preserving gradient norms,  and enhancing feature transferability of deep networks. We also broke down the pipeline of a few-shot learner and based on our findings, we established three augmentations strategies that aid in minimizing overfitting and increasing data diversity. Our findings and empirical results confirm that a DBT regularized model is beneficial to few-shot classification and meta-learning in general.

{\small
\bibliographystyle{ieee}
\bibliography{main}
}

\clearpage

\begin{center}
\textbf{\large Supplementary Materials}
\end{center}
\setcounter{page}{1}

\section{VC dimension and sample complexity}
The Vapnik–Chervonenkis (VC) dimension is a measure of the capacity (complexity, expressive power, richness, or flexibility) of a set of functions. In our setting we focus on neural networks where all the weights are of  low discriminant rank (LDR)  such as the Toeplitz-like, Hankel-like, Vandermonde-like, and Cauchy-like matrices.

\subsection{Bounding VC dimension} 
{\bf Theorem 1} For input $x \in \mathcal{X}$ and parameter $\theta \in \mathbb{R}^W$, let $f(x,\theta)$ denote the output of the network. let $\mathcal{F}$ be the class of functions $\{ x \xrightarrow{}f(x, \theta): \theta \in \mathbb R^W\}$. Let $\mathbf{W}_l$ be the number of parameters up to layer $l$ i.e the total number of parameters in layer (1,2,...,$l$). we define the depth effective path as:
\begin{equation}
\begin{aligned}
\bar L := \frac{1}{W}\sum_{l=1}^L W_{l}, 
    \end{aligned}
    \end{equation}
Then the total number of computations units is given as :
\begin{equation}
\begin{aligned}
\mathbf U := \sum_{l=0}^L n_{l}
    \end{aligned}
    \end{equation}

 Inline with works of \cite{Bartlett1998AlmostLV,Harvey2017NearlytightVB,Warren1968LowerBF,Anthony1999NeuralNL}.If k=1, corresponding to piece-wise linear networks, it can be shown that:
\begin{equation}
\begin{aligned}
\mathbf VCdim(sign \,\, \mathcal{F}) = O(\bar LW log(pU)) = O(LW log W).
    \end{aligned}
    \end{equation}

{\bf Lemma 1. } Let $p_q,..., p_m $ be polynomials of degree at most d in  $n \leq m$ variables, then we define:
\begin{equation}
\begin{aligned}
  K:= |\{ (sign(p_1(x)),..., (sign(p_m(x))): x \in \mathbb R^{n}\}|,
    \end{aligned}
    \end{equation}
i.e., if $K$ is the  number of possible sign vectors given by the polynomials, then $K \leq 2(2emd/n)^n$.  To partition the parameter space $\mathbb R^W $ for a fixed input $x_j$, the output $f(x,\theta)$ on each region in the partition implies $\mathcal{S} = {P_1,..., P_N}$ of the parameter $\mathbb R^{\mathbb W}$. 

Hence, we have:
\begin{equation}
\begin{aligned}
  K \leq  \sum _{j=1}^{N}|\{ (sign f(x_1,\theta)),...,sign f(x_m,\theta)):\theta \in P_j\}|,
    \end{aligned}
    \end{equation}

Hence from Lemma 1, we can show that by recursive construction, $\mathcal{S}_{L-1}$  is a partition of $\mathbb{R}^{W}$ such that for $\mathcal{S} \in \mathcal{S}_{L-1}$ \cite{thomas2019learning}. The network output for input $x_j$ is a fixed polynomial of $\theta \in \mathcal{S}$  which collectively gives: 
\begin{equation}
\begin{aligned}
\begin{split}
  K & \leq  \sum _{j=1}^{N}|\{ (sign f(x_1,\theta)),...,sign f(x_m,\theta)):\theta \in P_j\}| \\
     & \leq 2(\frac{2emkd_{L}}{W_L})^{W_L},
 \end{split}
\end{aligned}
\end{equation}

with the size of $ \mathcal{S}_{L-1}$ and equation (6) we get:
\begin{equation}
\begin{aligned}
  {K} \leq \prod_{l=1}^{L}2(\frac{2emkd_{L}}{W_L})^{W_L}
\end{aligned}
\end{equation}
We can take logarithm and apply Jensen’s inequality, with
$\bar W$ := $\sum_{l=1}^{L} W_l:$ 
\begin{equation}
\begin{aligned}
  {K} &\leq \prod_{l=1}^{L}2(\frac{2emn_lpd_{l}}{W_L})^{W_L} \\ 
  & log_{2}{K} \leq L + \sum_{l=1}^{L}W_{l}log_{2}2(\frac{2emn_lpd_{l}}{W_L}) \\
  & = L + \bar W \sum_{l=1}^{L} \frac{W_{l}}{\bar W}log_{2}2(\frac{2emn_lpd_{l}}{W_L})\\
  & \leq L + \bar W log_{2} \bigg(\sum_{l=1}^{L} \frac{W_{l}}{\bar W}\frac{2emn_lpd_{l}}{W_L}\bigg) \\
  &  = L +  \bar W log_{2} \frac{2emnp \sum_{l=1}^{L}n_ld_{l}}{\bar W}
\end{aligned}
\end{equation}
We bound $ \sum_{l=1}^{L}n_ld_{l}$ using the bound on $d_l$; since the  degree of an LDR matrix $d_l$  is at most:

\begin{equation}
\begin{aligned}
&  \sum_{j=0}^{l-1} n_{l}d_{l} \leq c_1k^{l-1} \sum_{j=0}^{l-1} n_{j}^{c_2} \\
& \leq  LUc_1k^{L-1}U^{c_2} \leq U^{c_2 + 2_k^{L}}
\end{aligned}
\end{equation}
where $c$ is a constant $L \leq U$ thus

\begin{equation}
\begin{aligned}
 log_{2}{K} \leq  L + \bar W log_{2} & \bigg(\frac{2c_1empU^{2 + c_2k^{L}}}{ \bar W}\bigg) \\ 
\end{aligned}
\end{equation}
To bound the VC-dimension, if VCdim(sign$ \, \mathcal{F}$) = $m$ there exists $m$ data points $x_1,...,x_m$ such that the output of the model can have $2^{n}$ sign patterns\cite{Vapnik2000TheNO}. The bound on $log_2${K} then implies:
\begin{equation}
\begin{aligned}
& VCdim(sign \, \mathcal{F})  \leq L \\
&   + \bar W log_{2} \bigg(\frac{2c_1epU^{2 + c_2k^{L}}VCdim(sign \, \mathcal{F}}{ \bar W}\bigg) \\ 
& = O(\bar{L}W log(pU) + \bar{L}LWlogk) 
\end{aligned}
\end{equation}
Hence completing the proof. Since the number of parameters $W$ is around the square root of the number of parameters of a network (e.g doubly block toeplitz based network) with unstructured layers, the sample complexity of an LDR network is much smaller than that of unstructured networks (e.g CNN) which is beneficial for deep networks.
\begin{figure}[t!]
\begin{center}

  \includegraphics[width=6.5cm, height=5.0cm ]{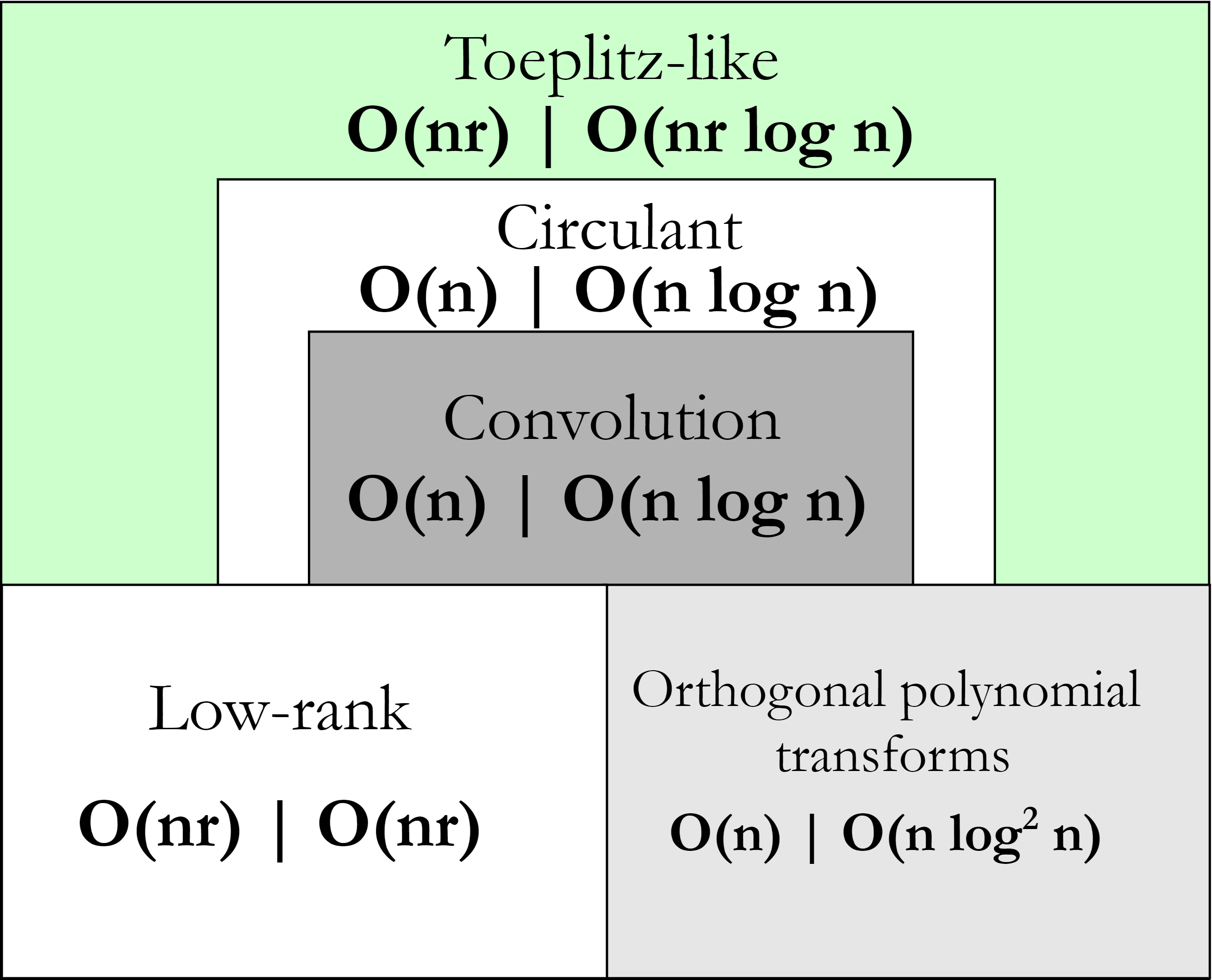}
\end{center}
\vspace{-.4cm}
  \caption{Captions for each class shows the storage cost and operation count for matrix-vector multiplication. Our proposed Toeplitz-like is of lowest rank. compared to circulant, standard convolutional filters, and orthogonal polynomial transforms.
} 

\vspace{-.5cm}
\end{figure}

\subsection{Space and time complexity.}
The  proposed  DBT-model has a time  complexity of  $ O(nr log n)$ and the small number of parameters also makes the network perform better with limited amount of training data which is crucial for few-shot learning \cite{Li2017LowRankDE}. We also ran tests on the  model backbone with two NVIDIA GeForce GTX 1080 Ti GPU and a batch size of 64. Table 1 reflects the accuracy performance and we see an overall model performance of about 4\%. 
Similar to \cite{Christiani2019FastLH}, the network used in our test consists of 4 convolutional layers, 1 fully-connected layer and one softmax layer.  Rectified linear units (ReLU) are used as the  activation  units. Images were cropped  to  24x24  and augmented with horizontal flips, rotation, and scaling trans-formations. We use an initial learning rate of 0.0001 and train for 800-400-100 epochs with their respective default weight decay. Our efficient DBT-based approach obtains a test error of 6.61\%,  compared to 5.26\% obtained by the conventional CNN model. At the same time, the DBT-based network is 4x more space efficient and 1.2x more time efficient than the conventional CNN-based model.

\subsection{Optimization setup.}
An SGD optimizer with a momentum  of  0.9  and  a  weight  decay  of  $2 \epsilon^{-5}$ was used for our setup.  We used a  learning rate initialized at 0.0001 with  a decay factor of 0.1 applied for all datasets. We trained over 100 epochs for miniImageNet and 200 epochs for both CIFAR-FS and 150 epochs for Stanford Dogs and Stanford Cats, respectively. 

\subsection{Architecture}
The works of \cite{Oreshkin2018TADAMTD,Lee2019MetaLearningWD,Tian2020RethinkingFI} used a ResNet12 as backbone for their model, we used a similar structure but replace the convolutional layer with a doubly-block toeplitz matrix the network consists of 4 residual blocks and  3 x 3 kernels. A 2x2 max-pooling layer is applied after each of the first 3 blocks; and a global average-pooling layer is on top of the fourth block to generate feature embeddings. Similar to \cite{Thomas2018LearningCT}, we used spectral regularization and changed the number of filters from (64, 128, 256, 512) to (64, 160, 320, 640). 

{\bf Usefulness of  DBT  regularization}
The DBT matrix represents a class of structured matrices  whose layers interact multiplicatively $(A^i,B^i)$ at $ O(nr log n)$ time as compared to convolutional layers that are linear and unstructured and are implemented in about $ O(n^2)$ time \cite{Li2017LowRankDE}. The generic term structured matrix refers to an $n$x$m$ matrix that can be described in fewer than $nm$ parameters and is capable of fast operation with at most double the displacement rank, which is far simpler for computations \cite{Li2017LowRankDE}. Hence, if $\mathcal{F}$ denotes a class of neural networks comprising of $L$ DBT layers, $W$ total parameters and piece-wise  linear activations, we can measure the complexity, expressive power, richness, or flexibility of $\mathcal{F}$ via a measure  referred to as the  Vapnik–Chervonenkis (VC) dimension \cite{Vapnik2000TheNO,Bartlett2003VapnikChervonenkisDO}.

For a simple classification problem of the form: $\{ x \rightarrow{sign}\;f(x): f \in \mathcal{F}\}$, the VC dimension $(VC_{dim})$ of the class is expressed as: 

\vspace{-.4cm}
\begin{equation}
\begin{aligned}
  VC_{dim}(sign \; \mathcal{ F}) = O(LWlogW).
    \end{aligned}
    \end{equation}
 \noindent   
 $\mathcal{V}C_{dim}(\cdot)$ matches the standard bound for unconstrained weight matrices \cite{Bartlett1998AlmostLV,Harvey2017NearlytightVB,thomas2019learning}.

\end{document}